\documentclass[conference]{IEEEtran}
\usepackage{times}

\usepackage[numbers]{natbib}
\usepackage{multicol}
\usepackage{graphicx}
\usepackage{amsmath}
\usepackage{amsfonts}
\usepackage{amssymb}
\usepackage{url}
\usepackage{multirow}
\usepackage{algorithm}
\usepackage{algorithmic}
\usepackage{color}

\usepackage{graphics} 
\usepackage{epsfig} 
\usepackage{times} 
\usepackage{amssymb}  
\usepackage{graphicx}
\usepackage{psfrag}
\usepackage{indentfirst}
\usepackage{subfigure}
\usepackage{url}
\usepackage{threeparttable}

\usepackage{comment}

\usepackage{algorithm}
\usepackage{algorithmic}

\usepackage{bm}

\usepackage{graphicx}
\usepackage{amsmath}
\usepackage{commath}
\usepackage{amsfonts}
\usepackage{amssymb}
\usepackage{url}
\usepackage{multirow}
\usepackage{algorithm}
\usepackage{algorithmic}

\usepackage{color}

\graphicspath{{img/}}

\begin{document}

\title{Towards Embodied Scene Description}

\author{Sinan Tan, Huaping Liu$^*$, Di Guo, Xinyu Zhang, Fuchun Sun\\Department of Computer Science and Technology, Tsinghua University\\ $^*$hpliu@tsinghua.edu.cn}

\maketitle

\begin{abstract}

Embodiment is an important characteristic for all intelligent agents (creatures and robots), while existing scene description tasks mainly focus on analyzing images passively and the semantic understanding of the scenario is separated from the interaction between the agent and the environment. In this work, we propose the \textit{Embodied Scene Description}, which exploits the embodiment ability of the agent to find an optimal viewpoint in its environment for scene description tasks. A learning framework with the paradigms of imitation learning and reinforcement learning is established to teach the intelligent agent to generate corresponding sensorimotor activities. The proposed framework is tested on both the AI2Thor dataset and a real world robotic platform demonstrating the effectiveness and extendability of the developed method.

\end{abstract}

\section{introduction}

\begin{figure}
\centering
\includegraphics[width=2.8in]{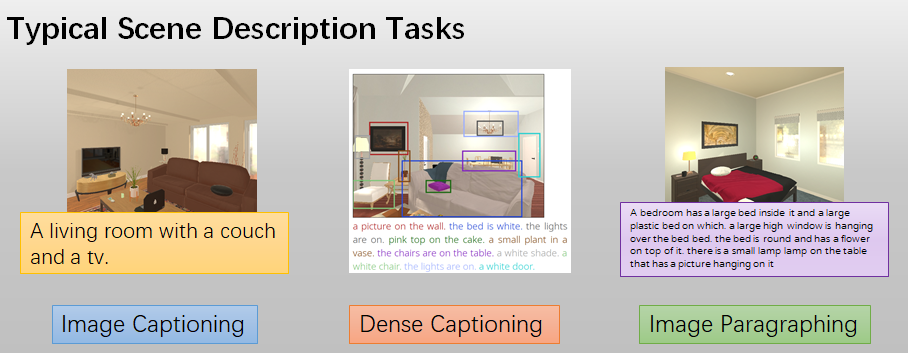}
\caption{Typical scene description tasks: \textit{Image Captioning}, \textit{Dense Captioning}, and \textit{Image Paragraphing}.}
\label{fig:Description_Tasks}
\end{figure}

When a visually impaired person enters a new room, he can easily take pictures of his surroundings using the smartphone and the built-in advanced computer vision modules are able to provide some scattered semantic information of these pictures. For example, the smartphone can detect certain classes of objects in the image with the help of an object detector and speak them out to the visually impaired person. However, such information is likely to make people confusing and uncomfortable due to its disorder and disorganization. A better way is to generate higher level semantic description such as natural language sentences or even paragraphs to describe the image. At present, great progress has been made in the areas of \textit{Image Captioning}\cite{vinyals2015show}\cite{xu2015show}, \textit{Dense Captioning}\cite{johnson2016densecap}, and \textit{Image Paragraphing} \cite{krause2017hierarchical}, and it has been becoming more and more mature with the booming of deep learning techniques\cite{he2016deep}. See Fig.\ref{fig:Description_Tasks} for some typical scene description tasks. It is believed that such semantic description will be an indispensable approach for the visually impaired people to perceive the environment\cite{seeingAI}. In this case, a further question arouses -- what is the next step?

In fact, no matter how accurate the semantic description is, it can only provide information that exists in the current image, but not tell the user what to do next. The semantic understanding of the scenario is separated from the interaction between the agent and the environment. When a visually impaired person enters a room, the first photo captured is likely to be a bare wall or window. At this time, he usually has to move the smartphone randomly expecting to capture a more meaningful image from another viewpoint. In this situation, it is more useful to tell him where to look next rather than just to provide him the vague description of the current scene (e.g. \textit{there is window on the wall}). On the other hand, a notorious problem of the semantic description is that it is very sensitive to the camera viewpoint\cite{park2019robust}. Although the content of the captured image may seem good, a deviation in camera viewpoint will lead to a totally wrong semantic description result. Under this circumstance, it is important to tell the visually impaired person how to adjust the position of the camera and even his body (e.g. move left, right, forwards, backwards) to get more meaningful and accurate scene description for the current scenario. Unfortunately, existing scene description work\cite{anderson2018bottom}\cite{johnson2016densecap}\cite{krause2017hierarchical} and the free APP software\cite{seeingAI} do not take this point into account. The reason is that all of them ignore the embodiment which is a very important characteristic of all intelligent agents (creatures and robots). The embodiment concept asserts that the intelligence emerges by interactions between the agent and the environment. Without embodiment, the semantic understanding of the scenario is separated from actions. It is a difficult problem which mainly involves the key issues of semantic scene description, description evaluation, and action instruction generation.

\begin{figure}
\centering
\includegraphics[width=3in]{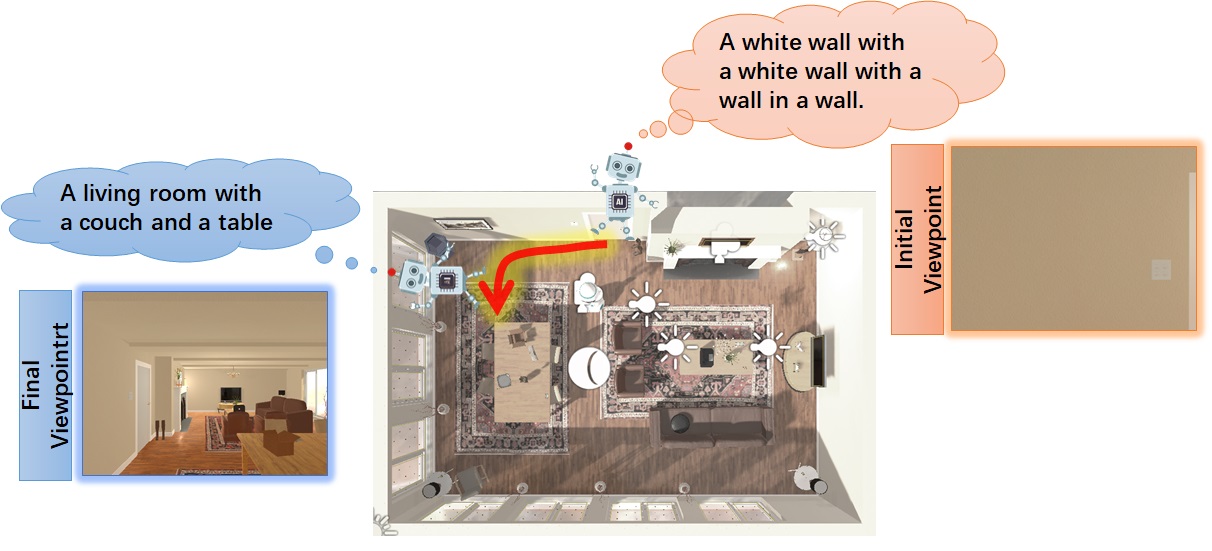}
\caption{An intuitive \textit{Embodied Scene Description} demonstration. Here we take the \textit{Image Captioning} as an example, while the idea is applicable to other tasks such as \textit{Dense Captioning}, \textit{Image Paragraphing}, etc. At first glance, the agent captures the initial image (rendered with \textit{pink}). Since this image is non-informative, the generated caption provides very limited information about the scene. However, the agent may explore the environment by itself to find a better viewpoint to capture a new image (rendered with \textit{blue}). The generated caption yields more informative and suitable results.}
\label{fig:demo}
\end{figure}

In this work, we propose the \textit{Embodied Scene Description} problem, which exploits the embodiment ability of the agent to find an optimal viewpoint in its environment for scene description tasks (e.g. \textit{Image Captioning}, \textit{Dense Captioning}, \textit{Image Paragraphing}, etc). The main idea is illustrated in Fig.\ref{fig:demo}. In addition to the visually impaired person, this problem is also extensively applicable to mobile robots. For example, it can facilitate the robot with many tasks such as actively exploring the unknown environment, quickly acquiring meaningful scene, and automatic photo taking.

To tackle this newly proposed problem, we establish a framework that makes use of existing image description models to guide the agent to explore an embodied environment. We encourage the agent to actively explore the environment and capture scenarios with good semantic description. It is noted that we consider the following two aspects when defining a good semantic description: (1) there should be sufficient visual objects detected in the scene and (2) these visual objects are able to compose a complete and reasonable semantic description. Since both the object detector and the semantic description may make mistakes, the combination of the two aspects is supposed to yield more reliable results. Having the definition of a good scene, we can build a learning framework with the paradigms of imitation learning and reinforcement learning to teach the intelligent agent to generate corresponding sensorimotor activities to explore the environment actively. It is worth noting that this work is different from a type of the embodied QA task\cite{gordon2018iqa}\cite{das2018embodied}, which is driven by finding answer to the question. In our work, the agent implements the task of environment exploration entirely with intrinsic motivation.

The main contributions are summarized as follows:

\begin{enumerate}
\item We propose a new framework for the \textit{Embodied Scene Description} problem, which exploits the embodiment characteristic of the intelligent agent to explore the environment to find the best viewpoint for scene description in an embodied environment.

\item We develop a learning framework with the paradigms of imitation learning and reinforcement learning to help the agent to acquire the intelligence to generate sensorimotor activities.


\item We testify the proposed method on AI2Thor dataset and evaluate its effectiveness using the quantitative and qualitative performance indexes.

\item We implement the proposed method on a robotic platform, which shows promising experimental results in real physical environment.

\end{enumerate}

\section{Related Work}
The deep learning methods have brought great success in many computer vision tasks such as object recognition\cite{he2016deep} and detection\cite{ren2015faster}. Moreover, many research studies have began to investigate a higher level task of semantic scene description with natural language. The proposed work focuses on the embodiment task of finding an optimal viewpoint for these scene description tasks.

Refs.\cite{donahue2015long}\cite{vinyals2015show} are some early-stage works that propose to use a combination of CNN and LSTM model to generate image captions. These image caption models are further improved by integrating different visual and semantic attention mechanisms\cite{xu2015show}\cite{you2016image}\cite{anderson2018bottom}. Due to the fact that information expressed in a single sentence is limited when describing an image\cite{song2018deterministic}, researchers begin to investigate some more complex models to bridge the gap between images and human language. Therefore, \textit{Dense Captioning}\cite{johnson2016densecap} is proposed, which describes an image with multiple sentences. Each sentence is corresponding to an area within a bounding box in the image. It is further improved by \textit{Image Paragraphing}\cite{krause2017hierarchical}, which is able to generate a long paragraph to describe an image instead of a single sentence. Ref.\cite{liang2017recurrent} proposes a better model for \textit{Image Paragraphing}, which utilizes the attention and copying mechanisms, as well as the adversarial training technique.

With the recent rapid development in computer vision and many traditional computer vision problems being addressed, the problem of embodied exploration has gradually emerged\cite{chen2019behavioral}. In the embodied exploration, an embodied agent actively explores the environment to have a better understanding of the scene\cite{li2019deep}. Contrary to traditional computer vision tasks, which mainly focus on analyzing static images passively, embodied exploration requires the agent both understands the content of the current image and takes proper actions accordingly to explore the environment. In most cases, the agent needs to make decisions based on observed image sequences instead of a single image\cite{sadeghi2019divis}.

Ref.\cite{zhu2017target} develops the target-driven visual navigation, where the agent tries to find an object that is given by an RGB image in an indoor scenario. The model is improved in \cite{ye2019gaple} by incorporating the semantic segmentation information. The embodied visual recognition task proposed in \cite{yang2019embodied} aims to address the problem of navigating in an embodied environment to find an object which might be occluded at first glance. Refs.\cite{jayaraman2018learning}\cite{ramakrishnan2019emergence} investigate the look-around behavior through active observation completion. Recently, language understanding and active vision are tightly coupled. In \cite{anderson2018vision}, the authors propose the task of visual-and-language navigation, where the agent is expected to follow the given language instructions, and use the collected vision information to navigate through the indoor scene. Refs.\cite{gordon2018iqa}\cite{das2018embodied} develop embodied question answering and interactive question answering tasks, where an agent is spawned at a random location in a 3D environment and explore to answer a given question. Such tasks have attracted many attentions from the computer vision communities\cite{yu2019multi}\cite{wijmans2019embodied}\cite{wu2019revisiting}\cite{das2018neural}. Although more and more work has taken the embodiment into consideration, the investigated tasks mainly focus on object search, scene recognition, and question answering. The problem of scene description in an embodied environment has not be investigated yet.

To solve the embodied perception problem, the deep reinforcement learning has become the most popular method for its ability to integrate the perception and action modules seamlessly. However, many scholars have pointed out that the end-to-end training for such complex tasks is rather difficult to converge\cite{jayaraman2018end}. To tackle this problem, some hybrid learning methods are proposed, such as sidekick policy learning which allows the agent to learn via an easier auxiliary task\cite{ramakrishnan2018sidekick}. In addition, some work prefers to use the imitation learning method\cite{wang2019reinforced}\cite{li2018oil} for pre-training and use reinforcement learning for fine-tuning\cite{haarnoja2018learning}. In this work, we resort to such methodology to solve the proposed \textit{embodied scene description} task.

\section{Problem Formulation}

The goal of this work is to develop a method to help the agent to rapidly find a proper viewpoint to capture a scene for generating the high-quality semantic scene description. Concretely speaking, we denote the image captured by the agent as $\bm{I}_t$ and the corresponding description as $\mathcal{U}(\bm{I}_t)$, where $t$ is the time instant. Please note that the operator $\mathcal{U}(\cdot)$ denotes the description generation procedure, which can be easily implemented by existing work, such as \textit{Image Captioning}, \textit{Dense Captioning}, \textit{Image Paragraphing}, and so on.

Fig.\ref{fig:demo} gives an intuitive introduction of the \textit{Embodied Image Description} problem. At time instant $t=0$, the captured $\bm{I}_0$ may contain \textit{wall} only and the produced caption \textit{a white wall with a white wall with a wall in a wall} is non-informative. Then the agent exploits its embodied capability to select an action to explore the room and get a new image. Such procedure is iterated until the agent captures an image containing plenty of objects and produces the informative caption \textit{A living room with a couch and a table}. The problem is therefore formulated as to develop an appropriate policy $\pi$ to help the agent to search a high-quality scene description about the scene. At each step $t$, the developed policy is used for the agent to take action $a_t$ to acquire the observed image $\bm{I}_t$.

Though our general idea is to learn action policies for an agent to locate a target scene in indoor environments using only visual inputs, the target scene is not specified by the user. This significantly differs from the work in \cite{zhu2017target}\cite{ye2019gaple} which requires a pre-specified target image.

\section{Navigation Model}
\begin{figure*}
	\centering
	\includegraphics[width=6in]{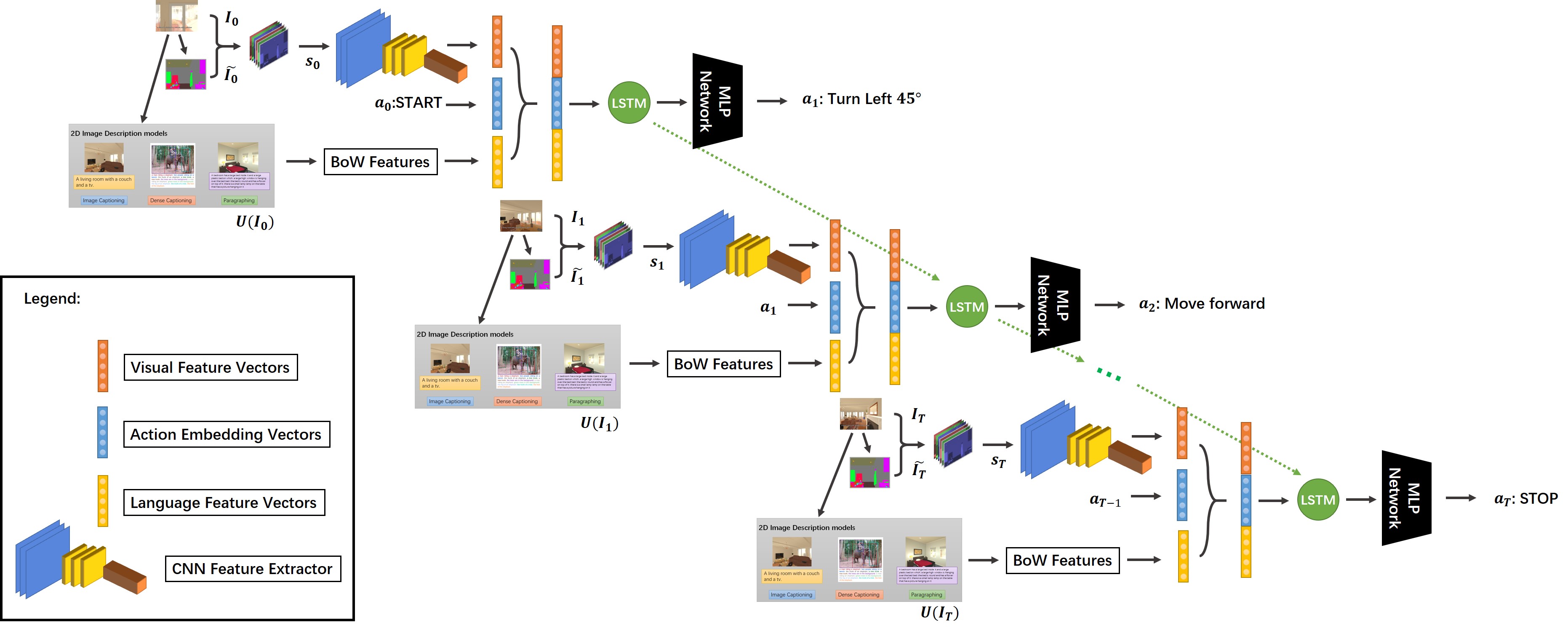}
	\caption{The proposed navigation model.}
	\label{fig:Network_Architecture}
\end{figure*}

\begin{figure}
	\centering
	\includegraphics[width=2.9in]{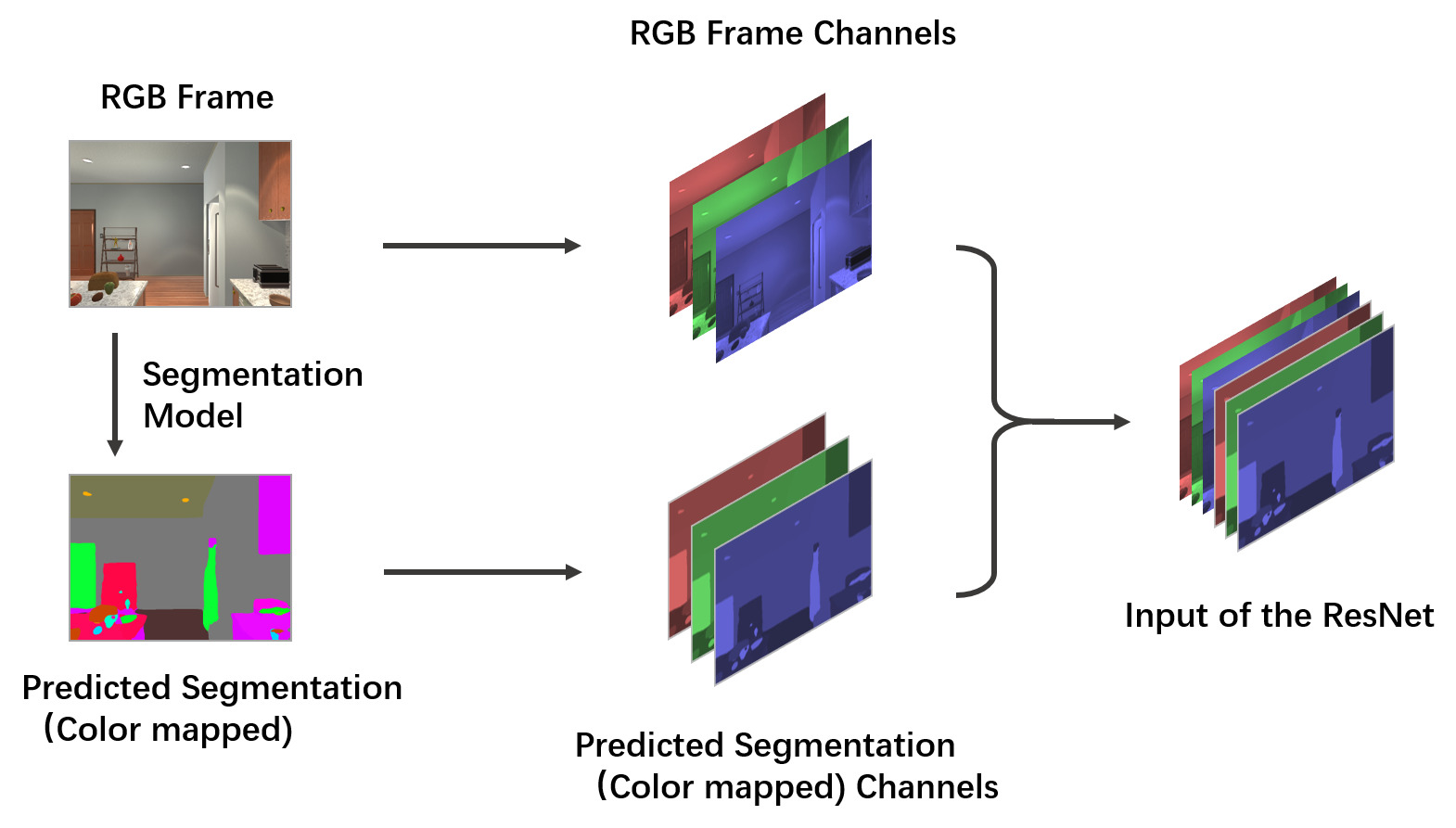}
	\caption{Demonstration of the visual input feed into the ResNet.}
	\label{fig:Input_Channels}
\end{figure}

The proposed navigation model is shown in Fig.\ref{fig:Network_Architecture}. The action the agent would take in one step can be relevant to all its previous actions and observations. Therefore, we model it using the LSTM network, which is very commonly used for sequence modeling \cite{vinyals2015show}\cite{xu2015show}.  With the learned policy, the agent is expected to take as few steps as possible to approach the target scene from a random starting position.

\subsection{State Representation}
In our implementation, we use a small ResNet-18 \cite{he2016deep} as a feature extractor, which is trained from scratch, jointly with the navigation model.

Since the image semantic segmentation results can improve the generalization performance\cite{ye2019gaple}, we also use the class segmentation map to help image description generation. To this end, we modify the number of input channels of the original ResNet-18 from 3 to 6. The added 3 channels are used to deal with the class segmentation map. Fig.\ref{fig:Input_Channels} shows how those 6 channels of the input fed to ResNet are generated. We use PSPNet \cite{zhao2017pyramid} to predict the class segmentation map for a given image.

Furthermore, since we need to train the model with thousands of images in one batch (100 scenes times approximately 30 steps at most for the demonstration trajectories of those scenes, which means about 3000 images in one batch), we shrink the original ResNet-18 to a smaller network with 10 convolution layers. There are 4 kinds of residue building blocks in the original ResNet18, and each of them is repeated twice, leading to 16 convolution layers in residue blocks with parameters (and there are another 2 convolution layers in ResNet18). We use each kind of those residue building blocks only once, yielding a model with only $18 - (16 - 8) = 10$ weighted layers. Besides, the output channels for all convolution layers are also halved (e.g. 512 output channels are shrunk to 256 channels for the final output layer).

Since the description result $\mathcal{U}(\bm{I}_t)$ can directly show how well the scene description model performs for the current frame, we extract the Bag-of-Words (BoW) feature $\bm{L}_{t}$ for all appeared words (after removing stop words) of the output of the 2D image understanding model.

Finally, we combine those multiple features to form the state representation. Denoting the class segmentation map of $\bm{I}_t$ as $\tilde{\bm{I}}_t$, the state vector can be represented as
\begin{equation}
\bm{s}_t = [ResNet(\bm{I}_t, \tilde{\bm{I}}_t); W_L \bm{L}_{t}]
\end{equation}
where $ResNet$ denotes the feature extraction module mentioned above. $W_L$ is a trainable parameter for language embedding.

\begin{figure}
	\centering
	\includegraphics[width=3in]{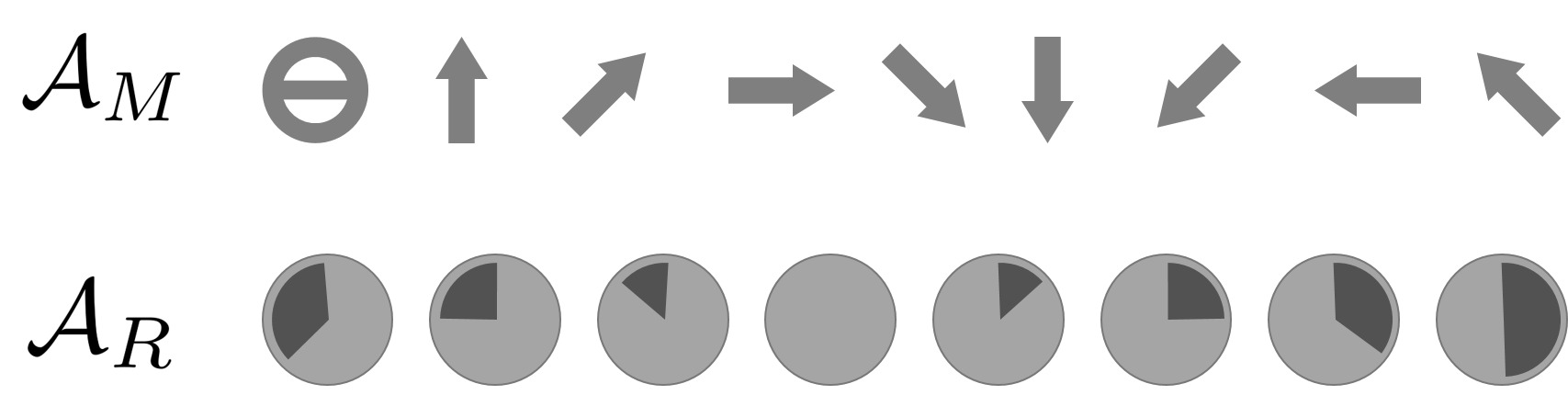}
	\caption{A representative action space. Please note that some actions (such as \textit{move left}) can be easily implemented in the simulation environment, but cannot be realized by some mobile agents, due to the non-holonomic constraints.}
	\label{fig:Action_Space}
\end{figure}

\subsection{Action Space}
As illustrated in Fig.\ref{fig:Action_Space}, for one step, we permit the agent to perform the following two kinds of discrete actions in the plane:
\begin{enumerate}
\item \textit{Move:} The agent can take nine basic actions which correspond to 8 directions and \textit{no move}. The move step is set to a fixed value $\Delta_m$ and the set of the movement actions is denoted as $\mathcal{A}_M$. In this work, we set $\Delta_m = 0.25$m.
\item \textit{Rotation:} The agent can rotate for a fixed interval of $\Delta_r$. In this work, we set $\Delta_r = 45^\circ$ and therefore the set of the rotation actions $\mathcal{A}_R$ contains 8 action atoms.
\end{enumerate}

For each step, the complete action space of the agent is $\mathcal{A} = \mathcal{A}_M \times \mathcal{A}_R$ and the agent is permitted to take the move actions firstly and the rotation action secondly. Please note that if the agent selects the action \textit{no move} from $\mathcal{A}_M$ and $0$ from $\mathcal{A}_R$, then the exploration is completed and the obtained image with description is reported as the final result.

In practical environment, the agent has motion limitation and may encounter obstacles or dead corner. Thus the selected action may not be realized. To solve this problem, the agent may use its sensors to detect the feasible region and construct the available action set $\mathcal{A}_t \subset \mathcal{A}$ for the $t$-th step.

\section{Matching between scene image and description}

\begin{figure}
	\centering
	\includegraphics[width=3in]{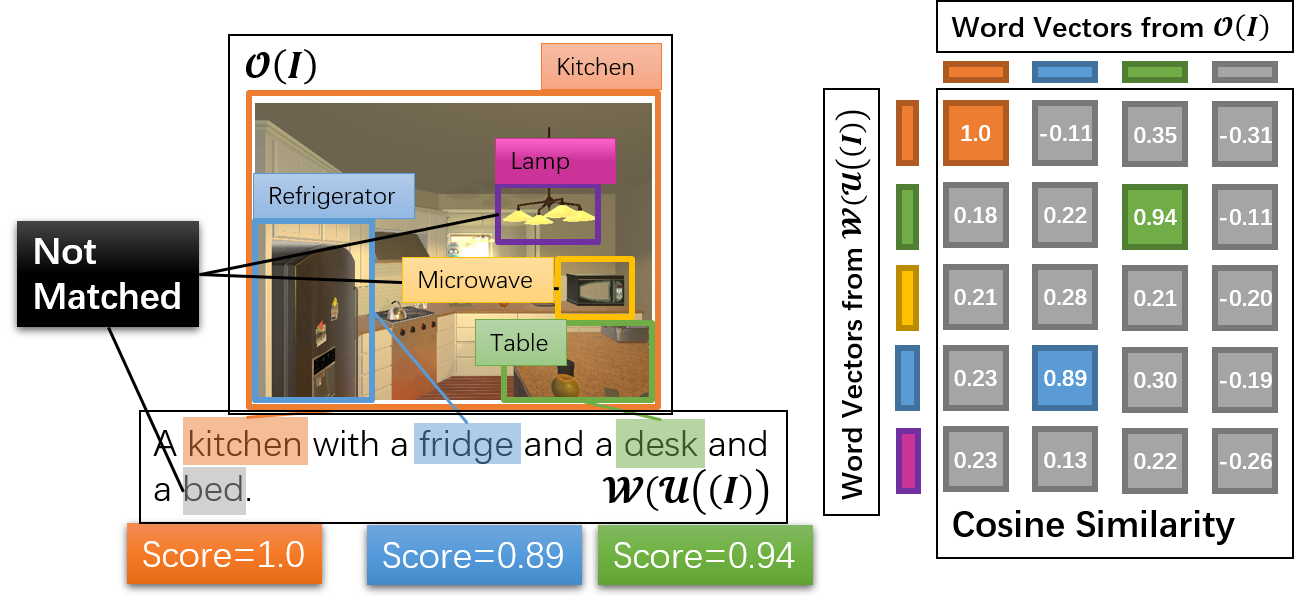}
	\caption{Demonstration of the proposed scoring function.}
	\label{fig:MatchingReward}
\end{figure}

\begin{figure}
	\centering
	\includegraphics[width=3in]{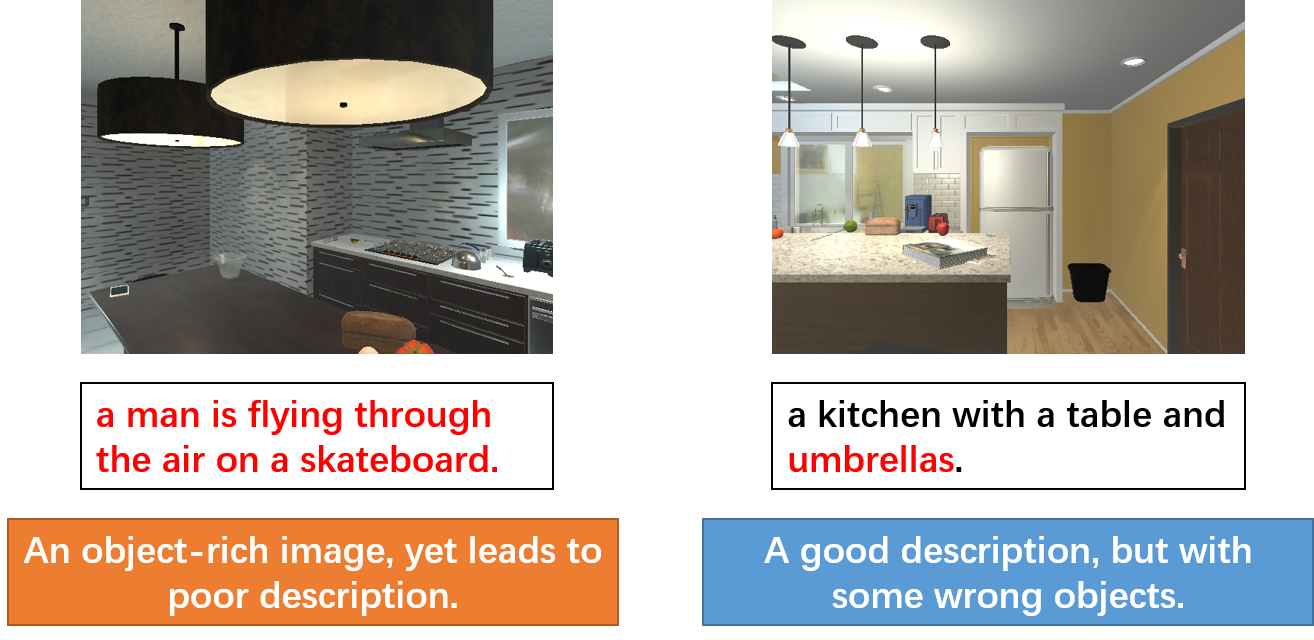}
	\caption{Some examples which show the mis-match between images and the corresponding description.}
	\label{fig:Image_Captioning_Failed_Cases}
\end{figure}

Since the goal of the navigation module is to guide the agent to find the scene which is good for both the image itself and the semantic description, we should design a matching score between the scene image and its description. This is indeed not a trivial task because the visual object detection results may contain noises and the image-text translator is far from perfectness. On one hand, an object-rich image is preferred but may lead to poor or non-informative description. On the other hand, a good description may include some wrong objects which do not appear in the image at all. See Fig.\ref{fig:Image_Captioning_Failed_Cases} for some examples.

To tackle this difficulty, we apply the off-the-shelf object detectors on the image $\bm{I}$ to find the visual objects, and extract the object nouns in the text description $\mathcal{U}(\bm{I})$. The matching score $score(\bm{I})$ is designed according to their connections. By matching those words with all of the detected objects for one image, we can quantitatively measure how ``good'' a viewpoint is.

We denote all appeared words in the category labels of all detected objects in the image $I$ as $\mathcal{O}(\bm{I}) = \{o_1, o_2, \cdots, o_n\}$, and all appeared noun words in the output of the description model $\mathcal{U}(\bm{I})$ as $\mathcal{W}(\mathcal{U}(\bm{I})) = \{w_1, w_2, \cdots, w_m\}$, where $n$ and $m$ are the numbers of the detected visual objects in the image $\bm{I}$ and the extracted noun words in the description $\mathcal{U}(\bm{I})$. Since the vocabularies adopted by the visual object detector and the semantic description may be different, the same object may be expressed by different words (such as \textit{desk} in the image and \textit{table} in the description). We resort to the Word2Vec \cite{mikolov2013distributed} embedding to semantically vectorize these words. For the object category label $o_i$ in $\mathcal{O}(\bm{I})$ and the noun word $w_j$ in $\mathcal{W}(\mathcal{U}(\bm{I}))$, we can define their similarly as
\[
R(o_i, w_j) = k(o_i, w_j) cos\langle o_i, w_j \rangle
\]
where $\cos \langle o_i, w_j \rangle $ is the cosine similarity between the word vectors of the two words. The value $k(o_i, w_j)$ is related to the confidence score of the word and the bounding box. Fig.\ref{fig:MatchingReward} is an intuitive demonstration of the matching-based score function.

The determination of the confidence $k(o_i,w_j)$ is dependent of the adopted description model. For example, if the adopted scene description task is \textit{Image Captioning} or \textit{Image Paragraphing}, since there is no specific information provided for the confidence score by the image description model, we just set $k(o_i, w_j) = 1$. For \textit{Dense Captioning} task, we have the confidence score and bounding box provided by the dense captioning model, therefore we can set $k(o_i, w_j) = IoU({BB}(o_i), {BB}({w_j}))\cdot C(w_j)$, where $BB(\cdot)$ is the corresponding bounding box and $C(\cdot)$ is the confidence score for the bounding box provided by the dense captioning model.

Based on the definition of $R(o_i, w_j)$, we can formulate the calculation of the similarity $sim(\bm{I}, U(\bm{I}))$ as the maximum matching problem between the sets $\mathcal{O}(\bm{I})$ and $\mathcal{W}(\mathcal{U}(\bm{I}))$. Such a problem can be easily solved using the Hungarian algorithm. Please note that this similarity value is normalized to [0,1].

Finally, we combine the similarity between image-description and the richness of objects to define the following viewpoint scoring function:
\begin{equation}
score(\bm{I}) = sim(\bm{I}, \mathcal{U}(\bm{I})) + \lambda \frac{ |\mathcal{O}(\bm{I})|}{N}
\label{eq:score}
\end{equation}
where $\lambda$ is a penalty parameter; the symbol $|\cdot|$ denotes the number of atoms in a set and $N$ represents the number of all possible objects. The second term encourages the agent to search the object-rich scene. It is very useful to some description tasks such as \textit{Image Captioning}, which usually contains few words.

\section{Learning for Embodied Scene Description}

A natural method to train the model presented in the previous section is the reinforcement learning algorithm. However, training such a complex model using end-to-end reinforcement learning from scratch is very hard to converge\cite{ramakrishnan2018sidekick}. Therefore, we first use demonstrations to develop imitation learning method to train the embodied scene description model from scratch, and then fine-tune this model with reinforcement learning. Such methodology has been extensively used for several difficult tasks\cite{das2018embodied}.

\subsection{Imitation Learning}

\begin{figure}
	\centering
	\includegraphics[width=3in]{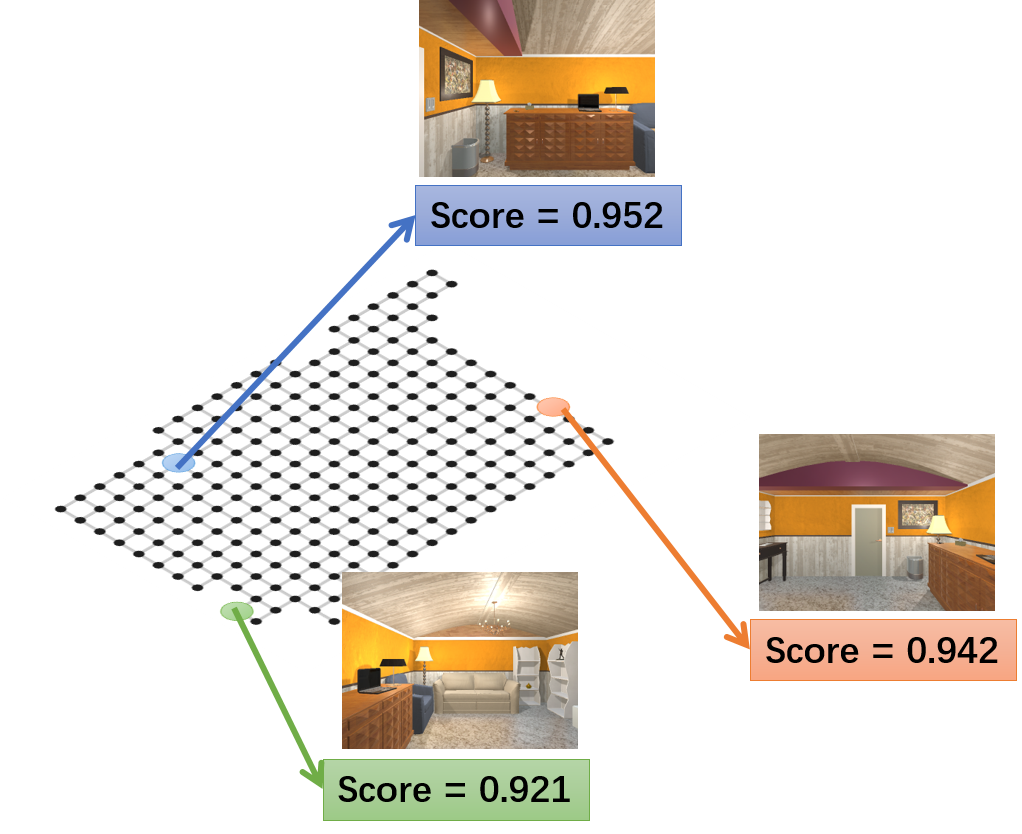}
	\caption{Demonstration of selecting the target locations.}
	\label{fig:GeneratedTargetPoints}
\end{figure}

\begin{figure}
	\centering
	\includegraphics[width=3in]{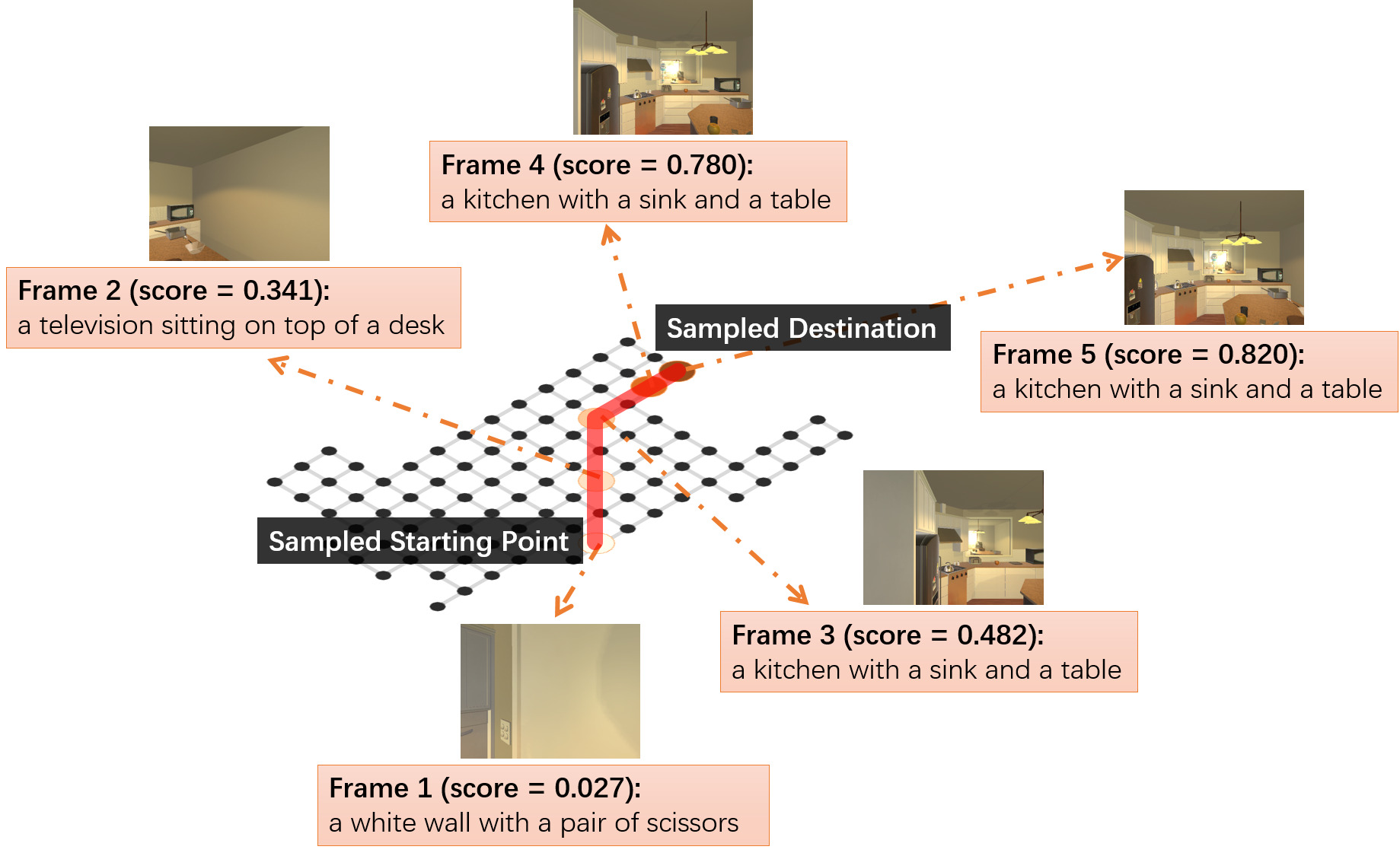}
	\caption{Demonstration of the generated shortest path.}
	\label{fig:GeneratedShortestPath}
\end{figure}

\begin{figure*}
	\centering
	\includegraphics[width=6in]{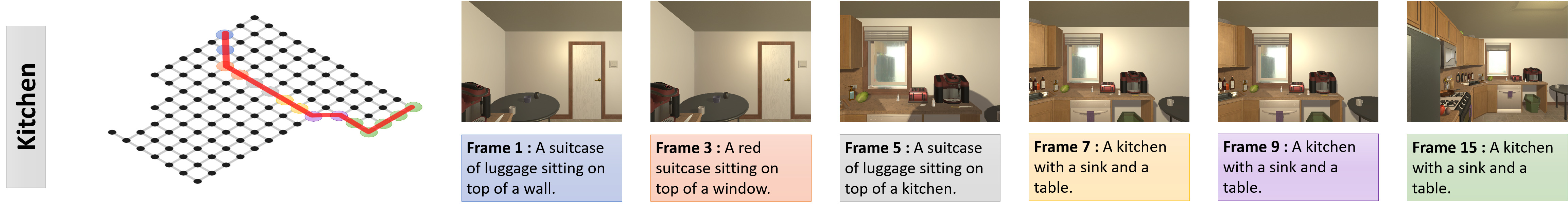}
	\includegraphics[width=6in]{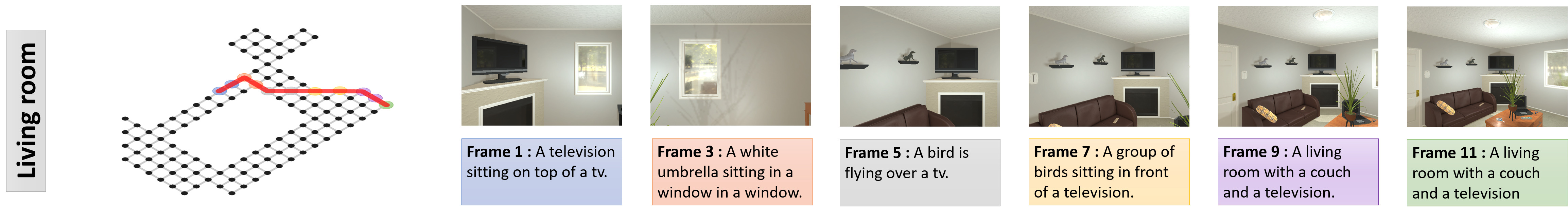}
	\includegraphics[width=6in]{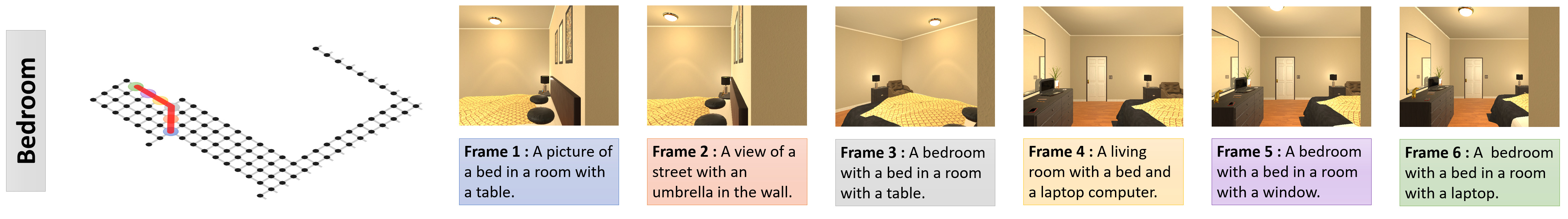}
	\includegraphics[width=6in]{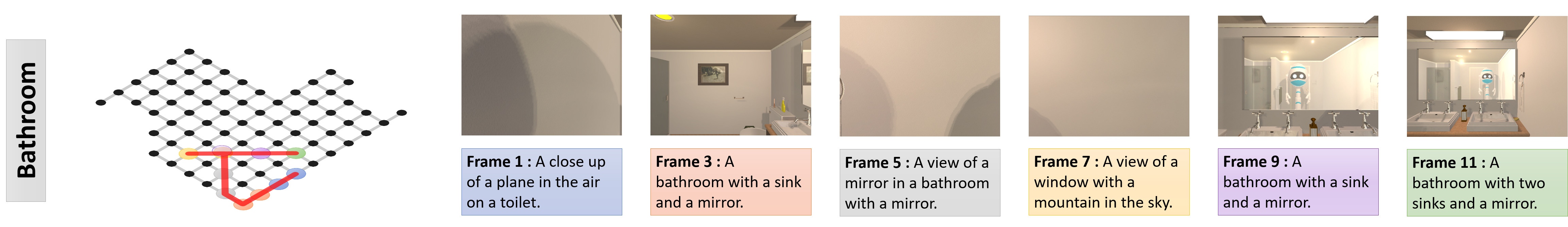}
	\caption{Four representative examples for the scenes \textit{Living Room}, \textit{Kitchen}, \textit{Bedroom} and \textit{Bathroom}. The trajectories of the agent are shown as red curves in the left panel.}
	\label{fig:Sample_Success}
\end{figure*}

The goal of imitation learning for sequential prediction problems is to train the agent to mimic expert behavior for some tasks. To develop the imitation learning algorithm, we have to annotate some scenes with the pre-trained caption model and generate demonstrations for the agent. Therefore, for a specific scene $\mathcal{S}$, we discretize it with grids of a fixed size of $\Delta_m$, and fixed angle $\Delta_r$ as is stated in the \textit{Action Space} section. For each possible position $(x,y)$ with rotation $\phi$, the corresponding viewpoint can be represented as the tuple $(x, y, \phi)$. We denote all these possible discrete viewpoints in the given scene $\mathcal{S}$ as $\mathcal{S}_D$. With some abuse of notation, we use
$score(x,y,\phi)$ to represent the score of the image which is captured at this viewpoint.

To produce the demonstration trajectories for a scene, we first find a special viewpoint $(x^*, y^*, \phi^*)$:
\begin{equation}
(x^*, y^*, \phi^*) = \mathop {\arg \max }\limits_{(x,y,\phi)\in \mathcal{S}_D} {score(x,y,\phi)}
\end{equation}
which achieves the highest score $s_{max} = score(x^*, y^*, \phi^*)$. Then we randomly sample one item from the set of candidate locations of which the score is in the interval of $[\gamma s_{max}, s_{max}]$ as the target location (demonstrated in Fig.\ref{fig:GeneratedTargetPoints}). The parameter $\gamma$ is set to 0.95 to prevent over-fitting. Finally, the shortest path between one randomly selected initial point and the target point, demonstrated in Fig.\ref{fig:GeneratedShortestPath}, can be obtained using the all-pairs shortest path table generated by the Floyd-Warshall algorithm\cite{hougardy2010floyd}. This path is used as the demonstration trajectory. Using the multiple demonstrations from various scenes, we can develop supervised imitation learning to train the feature extractor and the navigation model together. The loss function is defined as follows
\begin{equation}
\mathcal{L}_{\theta} = \sum_{k=1}^{K}{\sum_{t=1}^{T_k}{-\log{ \pi_\theta(\hat{a}_{k, t} | \hat{s}_{k, 0}, \hat{a}_{k, 0}, \hat{s}_{k, 1}, \hat{a}_{k, 1}, \cdots, \hat{s}_{k, t}) }}},
\end{equation}
where $K$ is the number of demonstration trajectories used for training in one batch, $T_k$ is the length of the $k$-th trajectory, $\hat{s}_{k, t}$ and $\hat{a}_{k, t}$ are the annotated observation and action, and $\theta$ denotes all of the parameters to be optimized. During the training phase, we assume a map of the environment is available and give the agent access to information about the shortest paths to some targets.

\begin{table*}
    \caption{Performance comparison}
    \centering
    \begin{tabular}{lccccccccccc}
	\hline
                                        & NoS & $SoL^*$\tnote{1} & $SoL$ & BLEU-1 & BLEU-2 & BLEU-3 & BLEU-4 & Meteor & ROUGE\_L & CIDEr \\ \hline
    Random                              & 26.78 & 0.3015 & 0.3017 & 0.6176 & 0.5088 & 0.4252 & 0.3686 & 0.2598 & 0.6086 & 1.7135 \\ \hline
    IL (RGB)                            & 21.38 & 0.7398 & 0.7430 & 0.8471 & 0.7997 & 0.7537 & 0.7144 & 0.4633 & 0.8293 & 4.8382 \\ \hline
    IL (Segm.)                          & 19.76 & 0.7524 & 0.7525 & 0.8607 & 0.8113 & 0.7611 & 0.7154 & 0.4653 & 0.8368 & 4.6627 \\ \hline
    IL (RGB+Segm.)                      & 15.43 & 0.7777 &\textbf{0.7734} & 0.8741 & 0.8334 & 0.7902 & 0.7502 & 0.4910 & 0.8625 & 4.9376 \\ \hline
    RL (RGB+Segm.)                      & 18.38 & 0.4490 & 0.4531 & 0.7228 & 0.6399 & 0.5682 & 0.5139 & 0.3401 & 0.7106 & 2.8099\\ \hline
	IL+RL(RGB+Segm.)                    & 15.10 & \textbf{0.7813} & 0.7724 & \textbf{0.8752} & \textbf{0.8345} & \textbf{0.7906} & \textbf{0.7502} & \textbf{0.4913}  & \textbf{0.8626} & \textbf{4.9482} \\ \hline
	\end{tabular}
\begin{tablenotes}
     \item[1] * denotes the evaluations on the validation set.
   \end{tablenotes}
	\label{table:Result}
\end{table*}

\subsection{Fine-Tuning with Reinforcement Learning}

After pre-training the navigation model with imitation learning, we then try to further improve its performance using the REINFORCE algorithm. The key to fine-tune the model is to design the reward function for the generated trajectory. Generally speaking, we hope to get high-quality caption within a short period of time and therefore a score can be designed as $p_t = score(\bm{I}_t) -  \rho t$, where $\rho$ is used to balance the scales of the two terms and is set to 0.01.

According to the above definition, the immediate reward is set as the incremental of the score $r(s_t, a_t) = p_t - p_{t-1}$ and the cumulative reward which is used to fine-tune the model is as
\begin{equation}
	R(s_t, a_t) = r(s_t, a_t) + \sum_{t'=t+1}^{T} \beta^{t'-t} r(s_t, a_t),
\end{equation}
where the discounted parameter $\beta$ is set to 0.99 and $T$ is the prescribed maximum steps and is set to 40.

Based on the above-defined reward function, we use REINFORCE algorithm to fine-tune the policy network. To reduce the variance of the reward and improve the stability, we record the moving average of the reward and minus the reward by that moving average in practical training. We use SGD optimizer with a learning rate of $10^{-3}$.

\section{Experiment Results}

The proposed framework is able to generalized to various semantic description tasks such as \textit{Image Captioning}, \textit{Dense Captioning}, \textit{Image Paragraphing}, and so on. Considering that the \textit{Image Captioning} provides a single-sentence description, which is more intuitive and convenient for practical applications, we focus on the scene description task of \textit{Image Captioning} in this section for performance evaluations. The experimental results on the tasks of \textit{Dense Captioning} and \textit{Image Paragraphing} are illustrated in the supplementary materials.

\subsection{Dataset}


For there isn't any existing dataset for the proposed \textit{Embodied Scene Description} task, we generate a new dataset with the AI2Thor dataset. A pretrained caption model is used to generate captions for each scene from different viewpoints.

The AI2Thor dataset contains 120 scenarios belonging to four categories: \textit{Living Room}, \textit{Kitchen}, \textit{Bedroom} and \textit{Bathroom}. Each category has 30 rooms. For each category, we use 25 rooms as the training set, and 5 rooms as the validation/test set. The layout of the room is discretized with grids. For those rooms used as validation/test set, one fixed point in every $4 \times 4$ grid is regarded as a point in validation set and the rest 15 points belong to the test set. The image caption model proposed in \cite{anderson2018bottom} is adopted for its satisfying performance. The model is trained with the MSCOCO captioning dataset\cite{vinyals2016show}.


\begin{figure}[h]
	\centering
	\includegraphics[width=3in]{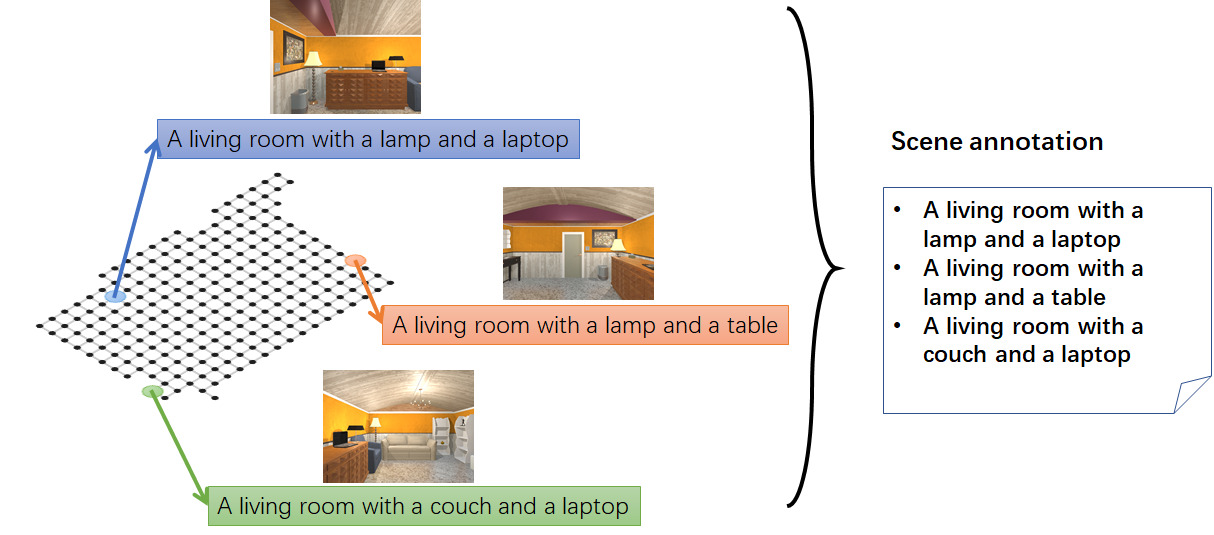}
	\caption{The annotation for the scene using selected three representative viewpoints.}
	\label{fig:scene_annotation}
\end{figure}

\subsection{Evaluation Metrics}
To evaluate the performance of the \textit{embodied scene description} task, we resort to the score function defined in Eq.(\ref{eq:score}) to calculate the score for each location in the room. Concretely speaking, for each scene, we select all of the locations whose scores are in the interval of $[\gamma s_{max}, s_{max}]$ and generate their corresponding captions. The generated captions are combined together to act as the ground truth annotation of the scene (Fig.\ref{fig:scene_annotation}). Based on this ground truth annotation, the following metrics are designed:


\begin{enumerate}
    \item \textit{Number of Steps (NoS):} The number of steps the agent takes before stopping.
    \item \textit{Score of the Last Image(SoL):} The score of the location which triggers the \textit{Stop} action.
    \item \textit{Natural Language Metrics:} With the generated ground truth annotation for each scene, metrics for natural language tasks can be used for evaluating the proposed task. We select several metrics including \textit{BLEU-1}, \textit{BLEU-2}, \textit{BLEU-3}, and \textit{BLEU-4} which are based on the $n$-gram precision\cite{papineni2002bleu}, \textit{Meteor} which considers the word-level alignment, \textit{ROUGE\_L} which is based on the longest common sub-sequence\cite{denkowski2014meteor}, and CIDEr\cite{vedantam2015cider}.

\end{enumerate}

\subsection{Result Analysis}
The performance of the proposed framework for the scene description of image captioning is illustrated in Table \ref{table:Result}. Comprehensive comparisons are conducted with several different settings. The full implementation of the proposed framework is denoted as \textit{IL+RL (RGB+Segm.)}. \textit{IL (RGB)} and \textit{IL (Segm.)} only use the imitation learning for the RGB image and segmentation map respectively. \textit{IL (RGB+Segm.)} uses both the RGB image and semantic segmentation map under the same imitation learning framework without reinforcement learning. We also investigate the performance of the reinforcement learning only framework which is trained from scratch and it is denoted as \textit{RL (RGB+Segm.)}. The baseline method is to randomly select each action.

\begin{figure*}
	\centering
	\includegraphics[width=6in]{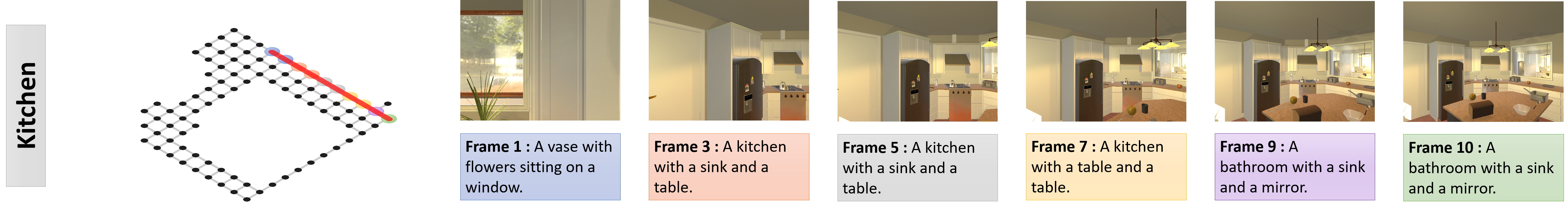}
	\caption{Failure case. Though an object-rich scene is finally discovered, the adopted caption model does not work well.}
	\label{fig:Sample_Failed_1}
\end{figure*}

The detailed results over all test samples are summarized in Table \ref{table:Result}, from which we have the following observations:
\begin{itemize}
\item \textit{The combination of IL+RL demonstrates good performance:} The results show that IL+RL method which contains both the pre-training process using imitation learning and fine-tuning process using reinforcement learning achieves the best performance according to all of the language-related metrics including BLEU, Meteor, ROUGE\_L and CIDEr. This verifies that the proposed method indeed helps the agent to find good viewpoints to get high-quality captions with an average of 15.10 steps that is satisfying among all the methods.

\item \textit{The RL only method works poorly.} The RL only method, though using the same information with IL+RL, yields very poor results, which is just better than the baseline random method. It demonstrates that the pre-training process using the imitation learning is helpful.

\item \textit{Both RGB and semantic segmentation information are important.} It can be seen that with the same imitation learning framework, using both the RGB image and semantic segmentation information has better performance than that with single modal information. The main reason is that RGB image is supposed to provide more details, while the segmentation map provides higher-level semantic information. In addition, IL(RGB) and IL(Segm.) methods take more steps before the stop action is triggered.
\end{itemize}
Although the obtained results are promising, we notice that the fine-tuning process using reinforcement learning only slightly improves the performance. This is in accordance with the results shown in existing literature \cite{das2018embodied}\cite{wang2019reinforced}. However, we believe the fine-tuning step could play more important roles when some more sophisticated strategies are utilized.

\subsection{Representative Examples}
In Fig.\ref{fig:Sample_Success}, we list four representative examples for different scenario categories. The agent is able to navigate in the room and finally find a good viewpoint to describe the scene. For example, for a living room which is illustrated in the second row of Fig.\ref{fig:Sample_Success}, the agent can only see the television initially, and then it continuously explores in the room until it reaches a position where a good view of the living room is obtained. For a bathroom which is illustrated in the last row of Fig.\ref{fig:Sample_Success}, the agent starts at a location where only the wall is visible. With the help of the navigation model, it gradually discovers the mirror and the sinks. Finally, it generates a description that contains major objects in the bathroom.

\begin{figure}
	\centering
	\includegraphics[width=3in]{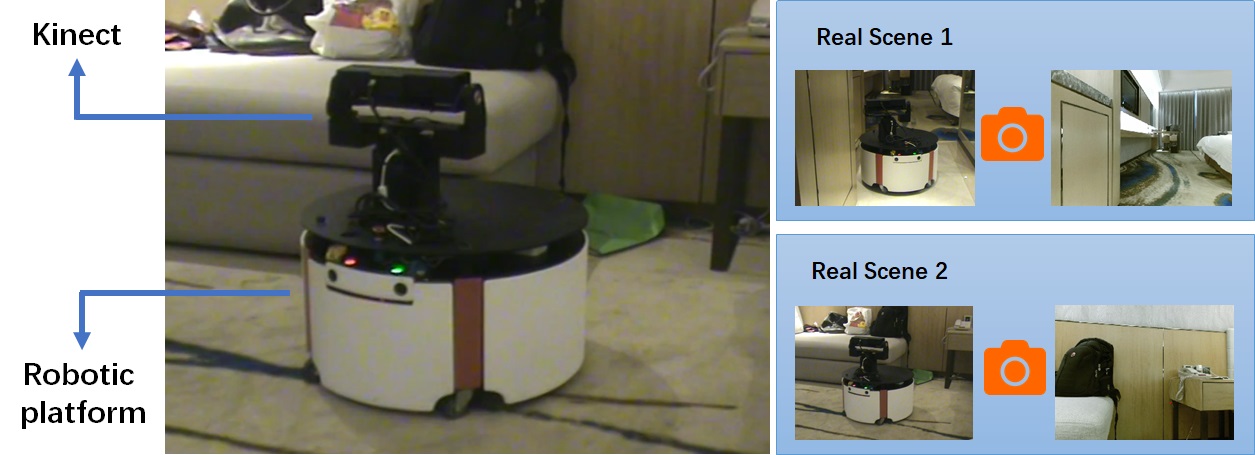}
	\caption{LEFT: The developed robotic platform. A Kinect is equipped on the top of it and we carefully adjust its position to ensure the optical center of the camera in Kinect is aligned with the center of the platform. RIGHT: Two representative real working scenes for the agent. For each scene, we show the the-third-person view (left) and the first-person view (right).}
	\label{fig:robot}
\end{figure}

It is noted that the performance of \textit{embodied scene description} is also strongly dependent of the scene description task (\textit{Image Captioning} in this task). In Fig.\ref{fig:Sample_Failed_1} we show a failure case. One possible reason is that the caption model generates improper caption for the scene even though the agent is actually find a good viewpoint for the scene description.

We also perform extensive experimental validation on some other typical scene description tasks such as \textit{Dense Captioning} and \textit{Image Paragraphing}. The results are shown in the Appendix and the video.

\subsection{Real-World Experiments}
Our model is trained on the AI2Thor dataset, rendered with Unity in high quality real-time realistic computer graphics, which makes the difference between the real world and simulation environments minimal. Therefore, it is possible for a simulation-to-real transfer and applying our model to real-world robots and scenes. The learned policy is able to provide action instructions to the robot.

\begin{figure}
	\centering
	\includegraphics[width=3in]{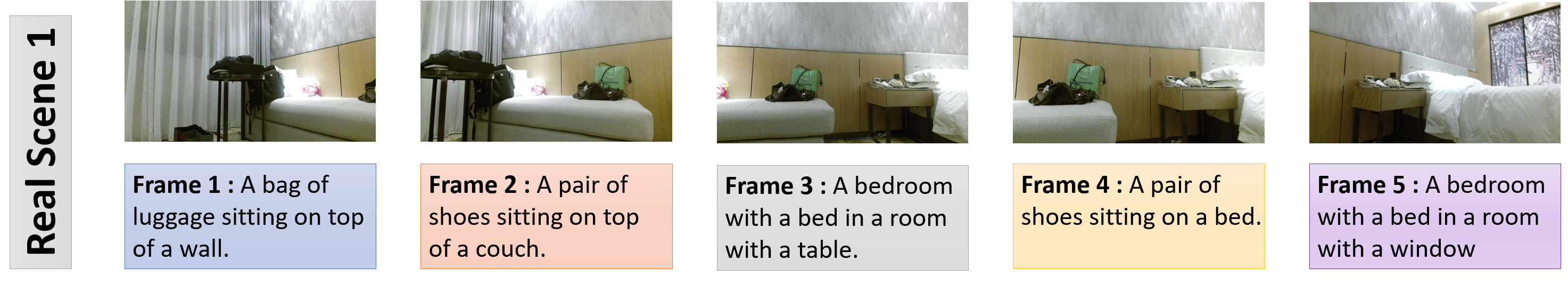}
	\caption{Real Scene 1: The robot turns around from the wall corner to the bed and finally correctly recognizes the bedroom scene.}
	\label{fig:Sample_Real_1}
\end{figure}

\begin{figure}
	\centering
	\includegraphics[width=2.8in]{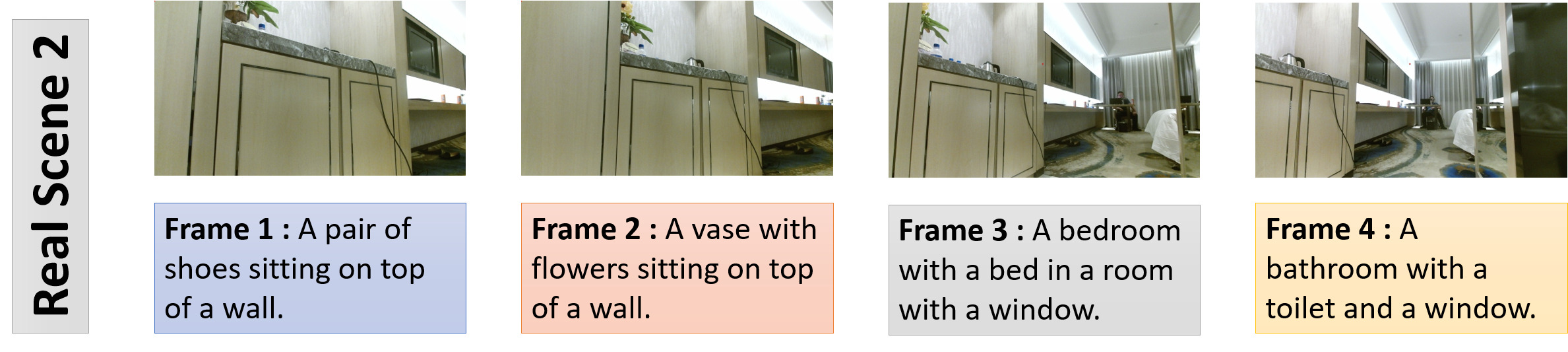}
	\caption{Real Scene 2: The robot adjusts its position to observe the room. But the caption model mistakenly recognizes the scene as a bathroom.}
	\label{fig:Sample_Real_2}
\end{figure}

As shown in Fig.\ref{fig:robot}, a mobile robot equipped with a Kinect camera is used in the real world experiment. The mobile robot is able to rotate 360 degrees around itself and move forwards/backwards flexibly, which allows for implementing actions in the action space. The Kinect camera is mounted on the top of the mobile robot and is used to collect egocentric images in real time. The robot is placed in an unseen hotel room for a simulation-to-real experiment. Although the layout of the room and the viewpoint of the camera are significantly different from those in the simulation environment, promising results are obtained to validate the effectiveness of the trained model. In Real Scene 1 (Fig.\ref{fig:Sample_Real_1}), the robot firstly faces to a corner of the room. With the generated instructions, the robot moves around the room until it recognizes that it is a bedroom scene. In Real Scene 2 (Fig.\ref{fig:Sample_Real_2}), the robot starts with a scene where it faces to the door of a cabinet and it is difficult to obtain much useful semantic description. Then, with generated action instructions, the robot adjusts its positions gradually and stops at a position where it can get a full view of the room. It reflects that the learned model is capable of transferring the semantic knowledge learned in simulation environment to real world environment. However, it can be seen that although the robot moves in a reasonable path, the captions generated are indeed wrong. It is because that the caption model used is not robust and accurate enough. More details can be found in the attached video.

\section{Conclusions}
In this work, we propose the new \textit{Embodied Scene Description} problem, in which the agent exploits its embodiment ability to find an optimal viewpoint in its environment for scene description tasks. A learning framework with the paradigms of imitation learning and reinforcement learning is established to teach the agent to generate corresponding sensorimotor activities. The trained model is evaluated in both the simulation and real world environment demonstrating that the agent is able to actively explore the environment for good scene description.

This work only takes one single frame into consideration for each step. Image sequences collected during the exploration process is believed to reveal more information for a better scene description. It will be also useful to leverage the attention, preference, and 3D relationship between objects to further actively understand the scenario. In the future, we plan to integrate this feature into smartphones and intelligent glasses, which are supposed to assist visually impaired person for a better living.

\bibliographystyle{IEEEtran}
\bibliography{references}

\end{document}